\documentclass[9pt, conference]{IEEEtran}
\IEEEoverridecommandlockouts

\usepackage{cite}
\usepackage{amsmath,amssymb,amsfonts}
\usepackage{algorithmic}
\usepackage{graphicx}
\usepackage{textcomp}
\usepackage{xcolor}
\usepackage{hyperref}
\usepackage{multirow}
\usepackage{multicol}
\usepackage{booktabs}
\usepackage{bbding}
\def\BibTeX{{\rm B\kern-.05em{\sc i\kern-.025em b}\kern-.08em
    T\kern-.1667em\lower.7ex\hbox{E}\kern-.125emX}}
\begin{document}

\title{Diagram Formalization Enhanced Multi-Modal Geometry Problem Solver\thanks{$\star$ Equal contribution. \quad \Envelope Corresponding author.\\Our model and dataset are available at \url{https://github.com/zezeze97/DFE-GPS}.}}


\author{\IEEEauthorblockN{Zeren Zhang$^{1 ,\star}$, Jo-Ku Cheng$^{1 ,\star}$, Jingyang Deng$^{1}$, Lu Tian$^{2}$, Jinwen Ma$^{1,}$\Envelope,\\Ziran Qin$^{3}$, Xiaokai Zhang$^{4}$, Na Zhu$^{4}$, Tuo Leng$^{4}$}
\IEEEauthorblockA{$^{1}$School of Mathematical Sciences, Peking University, Beijing 100871, China}
\IEEEauthorblockA{$^{2}$01.AI, Beijing, China}
\IEEEauthorblockA{$^{3}$School of Electronic, Information and Electrical Engineering, Shanghai Jiao Tong
University, Shanghai 200240, China}
\IEEEauthorblockA{$^{4}$School of Computer Engineering and Science, Shanghai University, Shanghai 200444, China}
\IEEEauthorblockA{\{eric\_zhang, chengruogu, jingyang\}@stu.pku.edu.cn, jwma@math.pku.edu.cn}}


\maketitle

\begin{abstract}
Mathematical reasoning remains an ongoing challenge for AI models, especially for geometry problems, which require both linguistic and visual signals. As the vision encoders of most MLLMs are trained on natural scenes, they often struggle to understand geometric diagrams, performing no better in geometry problem-solving than LLMs that only process text. This limitation is further amplified by the lack of effective methods for representing geometric relationships. To address these issues, we introduce the Diagram Formalization Enhanced Geometry Problem Solver (DFE-GPS), a new framework that integrates visual features, geometric formal language, and natural language representations. Specifically, we propose a novel synthetic data approach and construct a large-scale geometric dataset, SynthGeo228K, annotated with formal and natural language captions, designed to enhance the vision encoder to understand geometric structures better. Our framework improves MLLMs' ability to process geometric diagrams and extends their application to open-ended tasks on the formalgeo7k dataset.
\end{abstract}


\section{introduction}
As large language models (LLMs) improve in mathematical reasoning \cite{deepseekmath, qwen2-math, wizardmath}, interest grows in how multi-modal large language models (MLLMs) can get better at processing visual information that aids mathematical comprehension\cite{mathvista, mathv}. Solving geometry problems offers a practical way to evaluate MLLMs' mathematical reasoning abilities, as both visual diagrams and linguistic context are essential in this domain. This raises a fundamental question: Can MLLMs effectively understand geometric diagrams?

\begin{figure}[htbp]
    \centering
    \includegraphics[height=5cm]
    {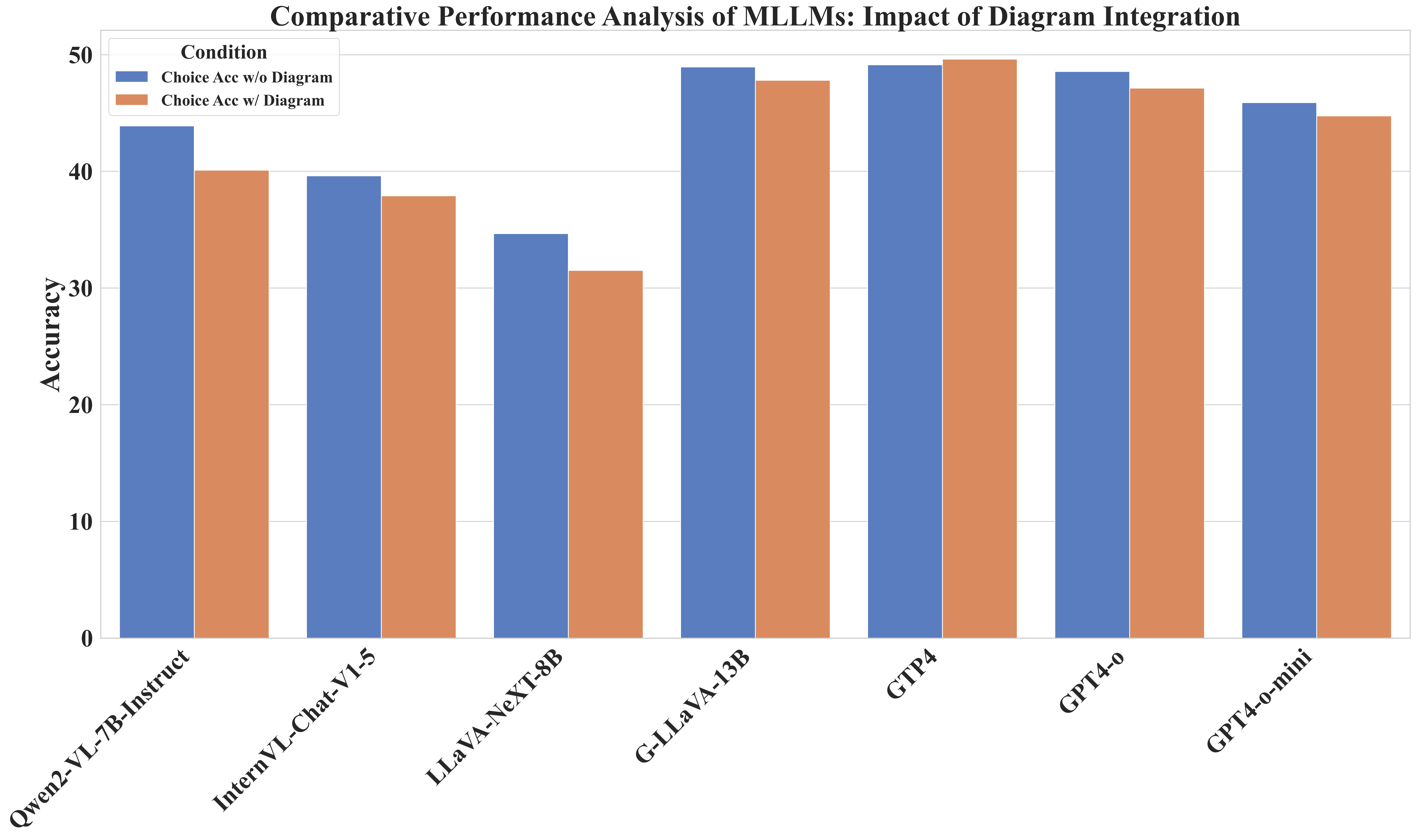}
    \vspace{-10pt}
    \caption{Comparative performance analysis of MLLMs: Impact of diagram integration}
    \vspace{-15pt}
    \label{fig:Comparative performance analysis of MLLMs}
\end{figure}

Studies indicate that MLLMs often struggle with geometric diagrams, showing no significant improvement over LLMs that only process textual information\cite{geoeval}. As shown in Fig. \ref{fig:Comparative performance analysis of MLLMs}, we further find that the geometry problem-solving accuracy of MLLMs decreases when geometric diagrams are introduced, compared to scenarios where images are absent or blank. Some LLMs perform even better than MLLMs at solving geometry problems, because MLLMs have difficulty interpreting these diagrams, which results in MLLMs extracting incorrect or irrelevant information from them\cite{gllava, hallusionbench}.

Several reasons may explain this discrepancy. Most LLaVA-like architectures \cite{llava, yi, deepseekvl} use vision encoders that are pre-trained on natural scenes, which significantly differ from geometric diagrams \cite{gllava}. Additionally, these models lack an effective methodology to represent geometric relationships.

\begin{figure*}[htbp]
    \centering
    \includegraphics[height=13cm]{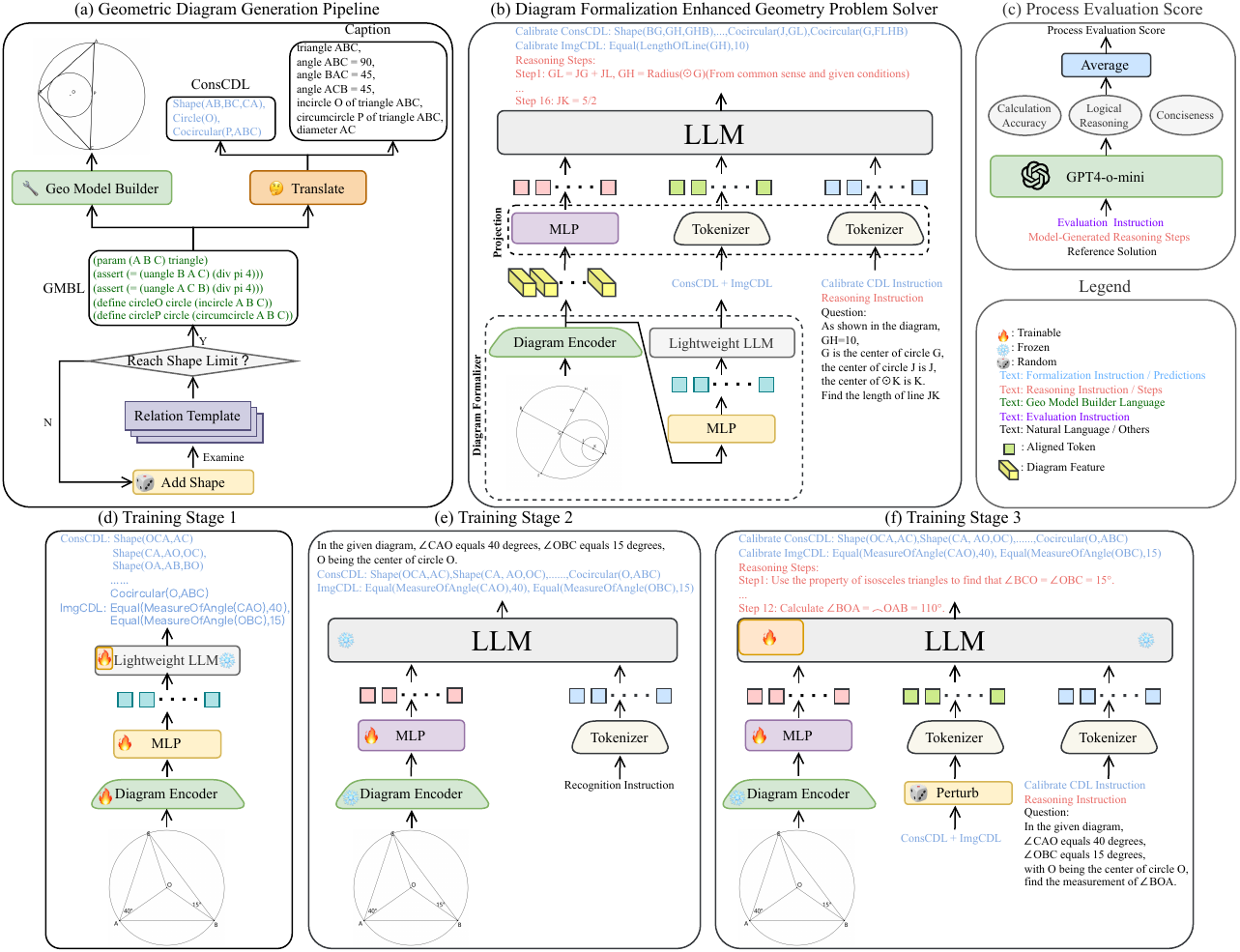}
    \vspace{-10pt}
    \caption{Our proposed geometric diagram generation pipeline (a), diagram formalization enhanced geometry problem solver (b) and process evaluation score (c). The three-stage training overview is illustrated in (d, e, f).}
    \label{fig:overall_pipeline}
    \vspace{-15pt}
\end{figure*}

To address these limitations, we introduce a new framework, the \textbf{D}iagram \textbf{F}ormalization \textbf{E}nhanced \textbf{G}eometry \textbf{P}roblem \textbf{S}olver (DFE-GPS), which incorporates a Diagram Formalizer — a model that leverages the formal language of these diagrams to improve the visual components of the model and boost the LLM's recognition of geometric structures. We outline a pre-training task for the vision encoder focused on diagram formalization and introduce a new pipeline for generating 228K large-scale geometric diagrams with both geometric formal language, and natural language descriptions from scratch utilizing geometry model builder \cite{geomodelbuilder}. This approach addresses the limitation of insufficient and misaligned diagrams in pre-training\cite{geogpt4v}. By employing geometric formal language, the LLM can effectively master properties and relationships of geometric elements such as points, lines, and triangles and generate human-readable, step-by-step solutions. Consequently, our model can better comprehend geometric diagrams. We summarize the contributions as follows:
\begin{enumerate}
    \item We introduce the DFE-GPS that integrates visual features, geometric formal language, and natural language, significantly enhancing geometric problem-solving as evidenced on the formalgeo7K test set.
    \item We propose a novel synthetic data approach combining geometric diagrams with formalized language, optimizing vision encoder pre-training for better feature extraction from geometric diagrams. Additionally, we release the SynthGeo228k dataset.
    \item We expand geometric problem-solving from multiple-choice questions to more challenging open-ended question answering, utilizing GPT-4o-mini to evaluate the step-wise problem-solving process with the Process Evaluation Score.
\end{enumerate}
\section{related work}
Geometry problem solving (GPS) has attracted substantial attention within the AI community. This field of study is divided into two GPS categories: single-modal, which relies solely on linguistic input, and multi-modal, which combines both vision and language.

\textbf{Single-modal GPS} Alpha Geometry \cite{alphageo} is a neuro-symbolic system that includes a symbolic deduction engine, achieving performance comparable to an average International Mathematical Olympiad (IMO) gold medalist. Similarly, FGeo-TP \cite{fgeoTP} utilizes the T5 language model to predict theorem sequences within its geometric formalized system FGPS, solving SAT-level geometric problems.

\textbf{Multi-modal GPS} requires a more complex data integration. This kind of research focuses primarily on generating theorem sequences to feed into a symbolic geometry solver. However, inconsistencies in dataset annotation standards have presented challenges in model training and performance evaluation \cite{geos, geoqa, unigeo, PGPSNet}.

InterGPS \cite{intergps} converts text and diagrams into formal language for theorem prediction, but relies on a pre-constructed theorem knowledge base for final results. Also, PGPSNet \cite{PGPSNet} uses a Convolutional Neural Network (CNN) and a Gated recurrent units (GRU)-based model to generate solution programs, which require an executor for outputs.

Recent advancements have also explored the use of MLLMs for GPS. G-LLaVA \cite{gllava} uses ChatGPT-3.5 to create an augmented dataset, Geo170K, and trains LLaVA without tuning the vision encoder. GeoGPT4V \cite{geogpt4v} uses GPT-4V to simplify complex problems into geometric QA pairs and generates corresponding images. However, it relies on the pre-existing questions from the dataset and cannot independently create diagrams.
\section{proposed approach}
This section begins with a review of the geometric formal language used in our approach, followed by an overview of the proposed geometric diagram generation pipeline. We then introduce the Diagram Formalization Enhanced Geometry Problem Solver (DFE-GPS) and detail the calculation of the Process Evaluation Score.

\subsection{Preliminary: Geometric Formal Language}
The theory of geometry formalization establishes a comprehensive framework for plane geometry. Formalgeo7k\cite{formalgeo} is annotated with Conditional Declaration Language (CDL), which encompasses construction CDL, text CDL, image CDL, and goal CDL. We utilize construction CDL (ConsCDL) and image CDL (ImgCDL) to represent geometric diagrams. ConsCDL conveys geometric structure information, including basic shapes, collinearity, and cocircularity, while ImgCDL offers geometric and algebraic relations, such as segment length and angle relationships.

\subsection{Geometric Diagram Generation}\label{subsec:geometric-diagram-generation}
Numerous methods exist for generating geometric diagrams, such as GeoGebra \cite{geogebra} and MATLAB \cite{matlab}, but these typically require human interaction and cannot produce a wide range of meaningful diagrams autonomously. In our geometric diagram generation process, illustration in Fig~\ref{fig:overall_pipeline}(a), we create templates expressed in Geometry Model Building Language \cite{geomodelbuilder} by adding new shapes. These templates, which range from single shapes to combinations of three with different geometric relationships, include basic elements like points and lines, as well as various triangles, polygons, and circles. When incorporating new elements, we analyze their geometric relationships, exploring scenarios such as circle lines intersections, angle congruences, and various configurations. Each template is translated into the formal language, ConsCDL, accompanied by detailed natural language captions specifying the shapes and their geometric elements. Our dataset, comprising 462 templates, provides a comprehensive exploration of basic geometry through the relationships between points, lines, and shapes. The Geometry Model Builder inherently allows for generating diverse diagrams from a single template by modifying point positions and orientations, with rotation being a common data augmentation method. Consequently, we have successfully generated over 228,000 geometric diagrams, constituting our SynthGeo228K dataset.

\subsection{Structure and Training Process of DFE-GPS}\label{subsec:DFE-GPS}
As shown in Fig.~\ref{fig:overall_pipeline}(b), our proposed DFE-GPS model integrates multiple modalities and comprises three main components: a Diagram Formalizer, a Projection module, and a LLM. Specifically, the LLM processes three types of inputs: diagram features $\mathcal{F}_{D}$ extracted by the Diagram Encoder, formal diagram language representations (ConsCDL and ImgCDL) produced by the Diagram Formalizer, and natural language inputs that include problem statements and instructions. The Projection module aligns this information within the LLM's semantic space, enabling effective integration of these diverse inputs. As a result, the LLM first calibrates the formal representations predicted by the Diagram Formalizer to gain a deeper understanding of the diagram and then generates reasoning steps for problem-solving. For implementation, we employ the pre-trained SigLIP~\cite{zhai2023sigmoid} as the Vision Encoder, Qwen2-0.5B-Instruct\cite{qwen2} as the Lightweight LLM, and Yi-1.5-Chat\cite{yi} (9B or 34B) as the primary LLM.

The training process consists of three stages, each centered on auto-regressive generation tasks. Let $\mathcal{T}_{in}$ denote the input text, $\mathcal{D}$ represent the input diagram, and $\mathcal{T}_{tar}$ be the target output. The training goal is defined by the following loss function:
\begin{equation}
\mathcal{L} = - \sum_{i=1}^{N}\log p[\mathcal{T}^{i}_{tar} | \mathcal{T}^{(<i)}_{tar}, \mathcal{T}_{in}, \mathcal{D}],
\end{equation}
where $p(\cdot)$ represents the multi-modal generative model, $\mathcal{T}^{i}_{tar}$ indicates the $i$-th token of the target sequence, and $N$ is the length of the target output. Fig.~\ref{fig:overall_pipeline} (d,e,f) illustrates the three-stage training process.

\textbf{Stage-1}: The first stage focuses on training the Diagram Formalizer module, aiming to generate formalized language descriptions that correspond to geometric diagrams. During this phase, the parameters of the Vision Encoder and part of the Lightweight LLM (through LoRA~\cite{lora}) are trainable to improve the model's ability to extract visual features.

\textbf{Stage-2}: The second stage centers on training the Projection modules to align vision features with the LLM's semantic space. This is achieved by generating natural language descriptions and formalized expressions for the geometric diagrams. During this stage, the parameters of the Diagram Encoder and the LLM are frozen, and only the MLP parameters linking visual features to the language model are trainable.

\textbf{Stage-3}: In the third stage, instruction fine-tuning enables the model to calibrate formalized diagram representations and solve problems. The input consists of geometric diagrams, formalized descriptions with random perturbations that simulate Diagram Formalizer errors, accompanied by problem text and calibration and reasoning instructions. The model learns to calibrate ConsCDL and ImgCDL, and then generate coherent natural language reasoning to solve the problem. During this stage, the parameters of the Diagram Encoder remain fixed, while the MLP and LLM parameters are trainable. Full parameter tuning is applied to the 9B model, whereas LoRA tuning is employed for the 34B model.

\subsection{Process Evaluation Score}
In addition to verifying the correctness of the final answers generated by the GPS, evaluating the reasoning process is even more critical. However, since our model generates solution processes in natural language, traditional formal methods used for symbol-based models\cite{geos, geoqa, unigeo} are not applicable. To address this challenge, we propose an evaluation method that leverages LLMs to assess the generated solution process. As illustrated in Fig.\ref{fig:overall_pipeline}(c), we input the model-generated reasoning steps and the reference solution into GPT-4o-mini, following the provision of tailored evaluation instructions. The model then reviews each reasoning step and evaluates the process based on three criteria: calculation accuracy, logical coherence, and conciseness. The average of these scores constitutes the Process Evaluation Score (PES).
\section{experimental results}
\subsection{Datasets and Implementation Details}
We develop two datasets from formalgeo7k, namely formalgeo-structure774k, focusing on geometric diagram formalization, and formalgeo-reasoning238k, enhanced via LLM-driven data augmentation. We also create the SynthGeo228k dataset using the Geometry Model Builder. In Stage 1, the Diagram Formalizer is trained for 4 epochs on both formalgeo-structure774k and SynthGeo228k (batch size 128, LoRA rank 16). Stage 2 involves 1 epoch of training on formalgeo-structure774k (batch size 256). In Stage 3, the 9B model undergoes full-parameter fine-tuning on formalgeo-reasoning238k (batch size 128), while the 34B model is fine-tuned using LoRA (rank 128, batch size 128). The entire process is accelerated using 8 NVIDIA A800 GPUs.

\subsection{Performance of Diagram Formalizer}
Table~\ref{tab:formalization performance} presents the performance results of the Diagram Formalizer on the formalgeo7k test set. The sentence-level accuracy metric assesses the correctness of individual compositional statements within the predicted CDL, whereas the full-expression accuracy metric evaluates the accuracy of the entire predicted CDL. Our Diagram Formalizer exhibits strong overall performance, with a significant improvement attributed to the incorporation of synthetic data. In contrast, excluding synthetic data leads to a considerable decline in performance, especially in terms of full-expression accuracy.

\begin{table}[htbp]
    \vspace{-5pt}
    \centering
    \caption{the performance of diagram formalizer and its variants on the formalgeo7k test set.}
    \vspace{-5pt}
    \renewcommand{\arraystretch}{0.8}    
    \begin{tabular}{ccccc}
        \toprule
        \multirow{2}{*}{Model} & \multicolumn{2}{c}{ConsCDL} & \multicolumn{2}{c}{ImgCDL} \\
        & Sentence & Full & Sentence & Full \\
        \midrule
        Diagram Formalizer & \textbf{90.25} & \textbf{72.29} & \textbf{92.88} & \textbf{84.38} \\
        w/o Synth-Data & 88.85 & 68.29 & 90.80 & 80.57 \\
        \bottomrule
    \end{tabular}
    \vspace{-5pt}
    \label{tab:formalization performance}
\end{table}

\subsection{Performance of Geometry Problem Solving}
Table~\ref{tab:main_expr} presents a comparative analysis of the problem-solving performance between state-of-the-art LMMs, MLLMs, and our proposed DFE-GPS on the formalgeo7k test set. Our evaluation significantly extends beyond multiple-choice questions to include the more demanding open-ended questions. In the multiple-choice mode, the model generates a solution and selects the correct answer from the provided options. In contrast, in the open-ended mode, the model independently generates a solution without any provided choices. We assess the models' performance by calculating the accuracy of the final answers across both modes. Furthermore, we utilize GPT-4o-mini to evaluate the correctness of the reasoning steps produced by the models, employing the PES.

Overall, the DFE-GPS-34B model exhibits outstanding performance across all metrics, achieving an accuracy of 82.38\% in the multiple-choice mode, 75.33\% in the open-ended mode, and a process evaluation score of 79.07. Although our smaller 9B model scores slightly lower, it still outperforms most other models. The increased difficulty of the open-ended mode leads to a decrease in accuracy across most models. Notably, DeepSeek-Prover-V1.5 \cite{deepseek-prover}, Yi-VL-6B\cite{yi}, and QwenVL \cite{qwenvl} fail to comply with instructions in the multiple-choice mode, often not selecting any of the provided options, resulting in accuracy rates lower than random guessing (25\%). Within the same model series, larger parameter models demonstrate superior performance, underscoring the benefits of increased model size. Furthermore, models specifically trained on mathematical data (Qwen2-Math\cite{qwen2-math}, DeepSeek-Math\cite{deepseekmath}, G-LLaVA\cite{gllava}) generally outperform general-purpose models. An exception is DeepSeek-Prover-V1.5, which, despite being trained under the LEAN \cite{lean} framework, performs poorly when confronted with natural language inputs.

\begin{table}[htbp]
    \vspace{-5pt}
    \centering
    \caption{performance comparison of formal enhanced gps against lmms and mllms on the formalgeo7k test set.}
    \vspace{-5pt}
    \renewcommand{\arraystretch}{0.8}    
    \begin{tabular}{cccc}
        \toprule
         \multirow{2}{*}{Model} & \multicolumn{2}{c}{Acc} & \multirow{2}{*}{PES} \\
          & Choice & Open-ended &  \\
         \midrule
         Llama-3.1-8B-Instruct\cite{llama3} & 26.38 & 16.57  & 48.56 \\
         Llama-3.1-70B-Instruct & 40.86 & 20.86  & 56.48 \\
         deepseek-math-7b-instruct\cite{deepseekmath} & 36.28 & 21.62 & 51.57 \\
         DeepSeek-Prover-V1.5-SFT\cite{deepseek-prover} & 19.24 & 34.10 & 66.14 \\
         DeepSeek-Prover-V1.5-RL & 20.38 & 34.67 & 65.50 \\
         Yi-1.5-9B-Chat\cite{yi}& 33.43 & 28.48 & 60.63 \\
         Yi-1.5-34B-Chat & 38.29 & 25.62 & 61.18 \\
         Qwen2-7B-Instruct\cite{qwen2} & 36.95 & 25.71 & 61.26 \\
         Qwen2-Math-7B-Instruct\cite{qwen2-math} & 56.38 & 42.95  & 71.12 \\
         Qwen2-Math-72B-Instruct & 57.71 & 49.81 & 76.89\\
         \midrule
         Yi-VL-6B\cite{yi} & 10.00 & 1.81 & 31.60 \\
         Yi-VL-34B & 31.88 &  3.24 & 37.50 \\
         QwenVL\cite{qwenvl} & 9.05 & 1.81 & 28.91 \\
         Qwen2-VL-7B-Instruct\cite{Qwen2-VL} & 40.10 & 22.19 & 61.84 \\
         Phi-3-vision-128k-instruct\cite{phi} & 39.24 & 14.00 & 49.47 \\
         InternVL-Chat-V1-5\cite{InternVL} & 37.90 & 16.29 & 51.47 \\
         LLaVA-NeXT-8B\cite{llavanext} & 31.52 & 10.19 & 48.83 \\
         LLaVA-NeXT-72B & 34.76 & 26.38 & 49.45 \\
         G-LLaVA-7B\cite{gllava} & 45.33 & 15.71 & 52.20 \\
         G-LLaVA-13B & 47.81 & 16.19 & 52.78 \\
         GPT-4-turbo\cite{gpt4} & 49.62 & 38.00 & 69.63 \\
         GPT-4o & 47.14 & 41.62 & 73.69 \\
         GPT-4o-mini & 44.76 & 40.00 & 69.23 \\
         \midrule
         DFE-GPS-9B & 77.05 & 68.67 & 76.00 \\
         DFE-GPS-34B & \textbf{82.38} & \textbf{75.33} & \textbf{79.07} \\
         \bottomrule
    \end{tabular}
    \vspace{-5pt}
    \label{tab:main_expr}
\end{table}

Beyond the GPT-4 series, MLLMs generally underperform compared to LLMs, likely due to their limited ability to effectively utilize diagrams. To explore this issue further, we conduct an experiment comparing the multiple-choice accuracy of MLLMs with and without the inclusion of images (Table~\ref{tab:impact of diagram}). While evaluating our model, we also control for the potential influence of formalized diagram language. The results indicate that for most MLLMs, incorporating diagrams often reduces accuracy in solving geometry problems, with QwenVL exhibiting a particularly notable decline. In contrast, our model demonstrates improved accuracy with the inclusion of diagrams, suggesting that our pre-trained Diagram Encoder effectively extracts features from diagrams and enhances overall performance.

\begin{table}[htbp]
    \vspace{-5pt}
    \centering
    \caption{effectiveness of mllms in utilizing diagrams for solving geometric problems.}
    \vspace{-5pt}
    \renewcommand{\arraystretch}{0.8}    
    \begin{tabular}{cccc}
    \toprule
        \multirow{2}{*}{Model} & \multicolumn{2}{c}{Choice Acc} & \multirow{2}{*}{$\Delta$} \\
                               & w/o D & w/ D & \\
    \midrule
         QwenVL~\cite{qwenvl} & 19.90 & 9.05 & \textcolor{red}{-10.85} \\
         Qwen2-VL-7B-Instruct\cite{Qwen2-VL} & 43.90 & 40.10 & \textcolor{red}{-3.80} \\
         Phi-3-vision-128k-instruct~\cite{phi} & 39.05 & 39.24 & 0.19 \\
         InternVL-Chat-V1-5~\cite{InternVL} & 39.62 & 37.90 & \textcolor{red}{-1.72} \\
         LLaVA-NeXT-8B~\cite{llavanext} &   34.67  & 31.52  &  \textcolor{red}{-3.15} \\
         G-LLaVA-13B~\cite{gllava} & 48.95 & 47.81 & \textcolor{red}{-1.14} \\
         GPT-4-turbo~\cite{gpt4}  & 49.14 & 49.62 & 0.48 \\
         GPT-4o & 48.57 & 47.14 & \textcolor{red}{-1.43} \\
         GPT-4o-mini & 45.90 & 44.76 & \textcolor{red}{-1.14} \\
         \midrule
         DFE-GPS-9B w/o CDL & \textbf{66.10} & \textbf{72.95} & \textbf{6.85} \\
     \bottomrule
    \end{tabular}
    \vspace{-5pt}
    \label{tab:impact of diagram}
\end{table}

\subsection{Ablation Study}
We examine the effects of omitting either Stage 1 or Stage 2 during training (Table~\ref{tab:ablation-1}) and observe significant performance drops, highlighting the essential role of our three-stage training process. Notably, Stage 1 pre-training with the Diagram Formalizer is more crucial than the process Stage 2.

\begin{table}[htbp]
    \centering
    \caption{evaluation of dfe-gps-9b across three training stages}
    \vspace{-5pt}
    \renewcommand{\arraystretch}{0.8}    
    \begin{tabular}{ccccc}
        \toprule
         \multicolumn{3}{c}{Training Stages} &  \multicolumn{2}{c}{Acc} \\
         Stage 1 & Stage 2 & Stage 3& Choice & Open-ended \\
         \midrule
         \checkmark & \checkmark & \checkmark & \textbf{77.05} & \textbf{68.67} \\
          \checkmark &  & \checkmark & 75.14 & 65.05 \\
            & \checkmark & \checkmark & 67.33 & 56.57 \\
         \bottomrule
    \end{tabular}
    \vspace{-5pt}
    \label{tab:ablation-1}
\end{table}

To investigate the impact of incorporating geometric formal language, we conduct experiments using three solution modes: (1) generating the solution directly without any CDL input; (2) incorporating the Diagram Formalizer's predicted CDL into our model to generate the solution; (3) building on the second approach by calibrating the predicted CDL before generating the solution. Table~\ref{tab:ablation-2} demonstrates a progressive improvement in accuracy across these approaches, suggesting that additional geometric formalization enhances the LLM's understanding of the diagrams, resulting in more effective problem-solving.

\begin{table}[htbp]
    \vspace{-5pt}
    \centering
    \caption{impact of different input and output on our 9b model's performance on the formalgeo7k test set.}
    \vspace{-5pt}
    \renewcommand{\arraystretch}{0.8}    
    \begin{tabular}{cc|cc|cc}
        \toprule
        \multicolumn{2}{c|}{Input} & \multicolumn{2}{|c|}{Output} & \multicolumn{2}{c}{Acc} \\
        D\&Q & Pred-CDL & Cali-CDL & Solution & Choice & Open-ended\\
        \midrule
        \checkmark &  &  & \checkmark & 72.95 & 64.76 \\   
        \checkmark & \checkmark &  & \checkmark & 76.19 & 67.81 \\  
        \checkmark & \checkmark & \checkmark & \checkmark & \textbf{77.05} & \textbf{68.67} \\
         \bottomrule
    \end{tabular}
    \vspace{-5pt}
    \label{tab:ablation-2}
\end{table}

\section{conclusion}


We have established the Diagram Formalization Enhanced Geometry Problem Solver (DFE-GPS) as a new and effective multi-modal large language model to leverage both geometric diagram and formalized information for solving geometric problems, along with a novel data generation pipeline for diagram formalization. The experimental results on the formalgeo7k demonstrate its significant improvements. Future work may explore reinforcement learning and tree search strategy to further enhance its performance.


\newpage
\bibliographystyle{IEEEtran}
\bibliography{refs}

\end{document}